\documentclass{jlcl}
\newcommand{\authornames}{Christoph Wick, Christian Reul, Frank Puppe}
\newcommand{\authorlastnames}{Wick, Reul, Puppe}
\newcommand{\articletitle}{Improving OCR Accuracy on Early Printed Books using Deep Convolutional Networks}

\newcommand{\shortarticletitle}{Deep Convolutional Networks for OCR}

\usepackage[TS1,T1]{fontenc}
\usepackage[utf8]{inputenc}
\usepackage{todonotes}
\usepackage{url}
\usepackage{hyperref}
\usepackage{multirow}
\usepackage{caption}
\usepackage{subcaption}

\def\longs{{\fontencoding{TS1}\selectfont s}}

\newcommand{\jlclvolume}{32}

\newcommand{\jlclnumber}{1}

\newcommand{\jlclyear}{2018}

\title{\articletitle}
\author{Christoph Wick\\
University of Würzburg\\
\texttt{christoph.wick@uni-wuerzburg.de} \and
Christian Reul\\
University of Würzburg\\
\texttt{christian.reul@uni-wuerzburg.de} \and
Frank Puppe\\
University of Würzburg\\
\texttt{frank.puppe@uni-wuerzburg.de}
}
 
\begin{document}

\setcounter{page}{1}
\thispagestyle{firstpage}

\authordata


\section*{Abstract}

This paper proposes a combination of a convolutional and a LSTM network to improve the accuracy of OCR on early printed books.
While the standard model of line based OCR uses a single LSTM layer, we utilize a CNN- and Pooling-Layer combination in advance of an LSTM layer.
Due to the higher amount of trainable parameters the performance of the network relies on a high amount of training examples to unleash its power.
Hereby, the error is reduced by a factor of up to 44\%, yielding a CER of 1\% and below.
To further improve the results we use a voting mechanism to achieve character error rates (CER) below $0.5\%$.
The runtime of the deep model for training and prediction of a book behaves very similar to a shallow network.

\section{Introduction}

The best OCR engines on early printed books like Tesseract (4.0 alpha)\footnote{\url{https://github.com/tesseract-ocr}} or OCRopus\footnote{\url{https://github.com/tmbdev/ocropy}} currently use Long Short Term Memory (LSTM) based models, which are a special kind of recurrent neural networks.
In order to achieve a low character error rate (CER) below e.g. 1\% or 2\% on early printed books these models must be trained individually for a specific book due to a high variability among different typefaces used.
Thereto, a certain amount of ground truth (GT), that is a pair of text line image and transcribed text in the case of OCRopus, must be manually labeled. The primary goal is to reduce the amount of labeled text lines to achieve a certain error rate.
A secondary goal is to continuously retrain the model, if more ground truth becomes available because e.g. all lines of a book are reviewed and corrected to achieve a final error rate of near 0\%. 
The default OCRopus implementation uses a shallow one layer bidirectional LSTM network combined with the CTC-Loss to predict a text sequence from the line image.
Since convolutional neural networks (CNN) showed an outstanding performance on many image processing tasks, see e.g. \cite{mane2017surveycnn}, our aim is to train a mixed CNN-LSTM network to increase the overall performance of OCR.
Therefore, we replace the default OCRopus implementation with an interface to TensorFlow\footnote{\url{https://www.tensorflow.org/}} in order to install our deep networks.
It is well known that voting the output of several different models improves the accuracy by a significant margin, which is why we use the proposed cross fold training approach of \cite{reul2017voting} to train five different models.
However, our simple voting method that is based on the ISRI analytic tools \citep{rice1996isri} only chooses the best sequence based on alignment and voting for the most frequent character instead of including the confidence values of each output into the voting mechanism.

The rest of the paper is structured as follows: Chapter 2 introduces and discusses related work on OCR on early printed books including deep models and voting.
The used data and the applied methods are described in detail in chapter 3.
In chapter 4 the results achieved on three early printed books are evaluated and discussed, before chapter 5 concludes the paper.

\section{Related Work}

This section lists related work concerning the application of CNN-LSTM hybrids in the areas of speech, vision, and text processing.
Furthermore, related work that show improvements on OCR using different voting algorithms are itemized.

\subsection{Combinations of CNN-LSTM}
Currently CNN-LSTM hybrids are used in a high diversity of fields to achieve state-of-the-art results.
The combination of those diverse network structures is promising because CNNs are suited for  hierarchical but location invariant tasks, whereas LSTMs are perfect at modelling temporal sequences.

For example, in the domain of speech processing \cite{sainath:2015:speech_cnn_lstm} use a combination of LSTMs, CNNs and Fully Connected Layer for an automatic speech recognition task, or else \cite{Trigeorgis:2016:Speech_emotion_recognition} train a CNN-LSTM for speech emotion recognition.
Another excellently suited area for CNN-LSTM networks is video processing, since CNN are perfect for handling images, and the video itself is a sequence of images.
For instance, in this area \cite{donahue:2015:LSTM-CNN-Visual-Description} use a CNN-LSTM as basic structure to automatically generate video descriptions or
\cite{Fan:2016:Video-Emotion-Recognition} use this structure for recognizing emotions in videos.

A combination of CNNs and LSTM in the field of text recognition was most recently proposed by \cite{breuel17hybridCNN-LSTM}.
The deep models yield superior results on the University of Washington Database III\footnote{\url{http://isis-data.science.uva.nl/events/dlia/datasets/uwash3.html}}, which consists of modern English prints with more than 95,000 text lines for training.
However, the effect of deep networks on historical books and using only a few hundreds of training lines was not considered, yet.

A very similar task to OCR is handwriting recognition e.g. by \cite{graves2009handwriting} or scene text recognition e.g. by \cite{shi:2017:scene-text}.
These applications usually act on contemporary data, which is why it is meaningful to include a language model or a dictionary to achieve a higher accuracy.
However, for early printed books, e.g. medieval books, general language models do not exist mostly due to variability in spelling.
Therefore, an OCR engine must be evaluated on how good it can predict each single character, whereas it is wanted that other text recognition systems also should be able to deal with misspelled words.

\subsection{Voting}

Using voting methods to improve the OCR results of different models was investigated by many different researchers, an overview is given by \cite{handley1998improving}.
Here, we list only a subset of the most important and the most recent work in the field of OCR on historical documents.

\cite{rice1996fifth} showed that voting the outputs of a variety of commercial OCR engines reduces the CER from values between 9.90\% and 1.17\% to 0.85\%.
The voting algorithm first aligns the outputs by using their Longest Common Substring algorithm \citep{rice1994algorithm} and afterwards applies a majority voting to determine the best character including a heuristics to break ties.

\cite{al2015combination} trained a LSTM network as voter of the aligned output of two different OCR engines.
A comparison using printings with German Fraktur and the University of Washington Database III the LSTM approach led to CERs (character error rates) around 0.40\%, while the ISRI voting tool achieved CERs around 2\%.
A major reason for this high improvements is that the LSTM-based voting algorithm learns some sort of dictionary.
Thus, the voter predicted a correct result although each individual voter failed.
However, this behaviour might not be desired since the method not only corrects OCR errors but also normalizes historical spellings.

Most recently, in \cite{reul2017voting} we showed that a cross-fold training procedure with subsequent confidence voting reduces the CER on several early printed books by a high amount of up to and over 50\%.
This voting procedure not only takes the actual predictions of a single voter into account, but also their confidence about each individual character.
Therefore, instead of a majority voting for the top-1 output class the best character is chosen for the top-N outputs.
This led to improvements by another 5\% to 10\% compared to the standard ISRI sequence voting approach.

\section{Material and Methods}
Our OCR pipeline is based on the OCRopus workflow, whose fundamental idea is to use a full text line as input and train an LSTM network to predict the ground truth character sequence of that line.
This is achieved by the usage of the CTC-Loss during training and a CTC-Decoder for prediction.
For a deeper insight in this pipeline, in this section, we first introduce the used datasets.
Afterwards, we present the basic ideas of the CTC-Algorithm and suggest our deeper models.
Then, we explain our training and evaluation procedure and finally, the used voting algorithm that further improves the results in a post-processing step.

\subsection{Datasets}
\begin{figure}[t]
\centering
\includegraphics[width=0.75\linewidth]{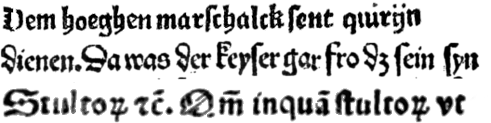}
\caption{Three example lines from each of the used books. From top to bottom: 1476, 1488, 1505.}
\label{fig:lines}
\end{figure}

For our experiments we use three different early printed books (see Table \ref{tab:books}).
Only lines from running text are used, whereas headings, marginalia, page numbers etc. are excluded, because these elements vary in line length, font size or their characters e.g. numbers are underrepresented.
Thus, the actual data is not affected by unwanted side effects resulting from these elements.
1505 represents an exception to that rule as we chose the extensive commentary lines instead, as they presented a bigger challenge due to very small. inter character distances and a higher degree of degradation
Fig. \ref{fig:lines} shows one line per book as example.

\begin{table}[t]
\centering
\caption{Books used for evaluation and their respective amount of ground truth lines available for training and validation.}
\label{tab:books}
\begin{tabular}{cccc}
\hline
\textbf{Year} & \textbf{Language} & \textbf{GT Train} & \textbf{GT Validation} \\
\hline
1476             & German             & 2000              & 1000 \\
1488             & German             & 3178              & 1000 \\
1505             & Latin              & 2289              & 1000 \\
\hline
\end{tabular}
\end{table}
1505 is an edition of the \textit{Ship of Fools} (\textit{Narrenschiff} by Sebastian Brant) and was digitized as part of an effort to support the \textit{Narragonien digital} project at the University of Würzburg\footnote{\url{http://kallimachos.de/kallimachos/index.php/Narragonien}}.
1488 was gathered during a case study of highly automated layout analysis \cite{reul2017dhl} and 1476 is part of the Early New High German Reference Corpus\footnote{\url{http://www.ruhr-uni-bochum.de/wegera/ref/index.htm}}.

\subsection{CTC-Algorithm}
\label{sec:ctc-algorithm}
The CTC algorithm is basically used to compute the probability of a label sequence given a probability distribution of all labels over time, which is equivalent to the width of the image in our case.
This is done by adding a blank label \texttt{\_} to the codec which allows the network to predict no or an empty character.
The algorithm then sums by dynamic programming the probability of all possible paths that match the regular expression \texttt{\_*A+\_*B+\_*C+\_*D+\_*}, where \texttt{A} through \texttt{D} are arbitrary labels.
Note that repetitions of labels must be separated by a blank.
During the training, the derivative of this probability is computed to update the weights of the network according to the gradient descent rules.

The resulting network output, which is the probability distribution of all labels at each point in time, must be decoded for application of the network.
The simplest decoder is the so-called CTC-Greedy-Decoder, which first detects the label with the highest probability for each point in time, thus computes the argmax.
Afterwards, according to the upper regular expression, all repeated labels are removed and finally all blanks are removed.
That means, repeated characters will be ignored if no blank label is interrupting, but blank labels are not necessary to separate different labels.

An adaption of the standard CTC-Algorithm matches paths using \texttt{\_*A\_*B\_*C\_*D\_*} as allowed representations.
The difference to the expression \texttt{\_*A+\_*B+\_*C+\_*D+\_*} is that repeated characters are not merged anymore, which makes it easier for the network to predict repeated labels.
Hence, the drawback of this method is, that a single character must be predicted at one sole position by the network, which restricts the freedom of the network by this means.

The standard CTC-Algorithm can be adapted by disabling the merge of repeated characters.

\subsection{Network architecture}
The standard OCRopus network uses a single hidden LSTM layer with 100 nodes.
We extend this structure by introducing convolutional and pooling layers before the default LSTM-Layer.
By default we use a convolutional kernel of size $3 \times 3$ with a stride of 1 and equal padding.
The network uses max pooling layers with a kernel size and stride of $2\times2$.
A stride or a kernel size of 2 in the first dimension, that is the time dimension, halves the width of the intermediate picture and therefore the number of LSTM operations.
On the one hand, this makes the network faster but on the other hand repeated characters might not get resolved, because they require an intermediate blank label.
To compensate this shortcoming, we add networks that either have a max-pooling size of $1 \times 2$ or disable the merge of repeated characters in the CTC-Loss and CTC-Decoder (compare Section \ref{sec:ctc-algorithm}).

Finally, we optionally add dropout with a factor of 0.5 to the last layer.
Since on average only half of the last LSTM nodes are active by using the proposed dropout, we add a last network with 200 LSTM nodes so that on average still 100 nodes are active during the training step.

Compare Table \ref{tab:networks} for an overview of all the tested network structures.
Here, Network 1 labels the default model which is trained by the standard OCRopus implementation, whereas all other networks are implemented by replacing the model of OCRopus using a TensorFlow backend.
Our source code is published at GitHub\footnote{\url{https://github.com/ChWick/ocropy/tree/tensorflow}}.

\begin{table}[t]
    \centering
    \caption{Overview of the tested models. CNN 40 3x3 indicates a convolution layer with 40 filters and a $3\times3$-Kernel size and always a stride of 1. Pool 2x2 adds a max-pooling layer with a kernel size of $2\times2$ and equal stride. LSTM 100 is a Bidirectional LSTM layer with 100 hidden nodes.}
    \label{tab:networks}
    \begin{tabularx}{\textwidth}{cX}
        \hline
        \textbf{Network ID} & \textbf{Structure} \\
        \hline
        1 & LSTM 100 \\
        2 & CNN 40 3x3, Pool 2x2, LSTM 100 \\
        3 & CNN 40 3x3, Pool 2x2, CNN 60 3x3, Pool 2x2, LSTM 100 \\
        4 & CNN 40 3x3, Pool 2x2, CNN 60 3x3, Pool 1x2, LSTM 100 \\
        5 & CNN 40 3x3, Pool 2x2, CNN 60 3x3, Pool 2x2, LSTM 100, CTC no merge repeated \\
        6 & CNN 40 3x3, Pool 2x2, CNN 60 3x3, Pool 2x2, LSTM 100, Dropout \\
        7 & CNN 40 3x3, Pool 2x2, CNN 60 3x3, Pool 2x2, LSTM 200, Dropout \\
        \hline
    \end{tabularx}
\end{table}

\subsection{Training and Evaluation}
Our TensorFlow implementation uses an Adam solver with an initial learning rate of 0.001.
The batch size of one is equal to the default OCRopus implementation, that means only one image is used per training step.
Greater values that would allow an efficient use of GPUs are beyond the focus of this paper.

First, each book in the dataset is split into an evaluation and a training set.
While the evaluation set size is fixed, the training set size is chosen as 60, 100, 150, 250, 500, and 1000.
Each training set is again divided into a 5-fold, where four parts are used for the actual training and one part for validation.
Therefore, for three books, six different numbers of lines in the training set, and seven network architectures, we train five models, respectively.
That comes to a total of $3 \cdot 6 \cdot 7 \cdot 5 = 630$ training experiments.
For each run, we compute the CER on the validation set every 1000 iterations to determine the best model of this fold.
This model is then evaluated on the initial evaluation dataset to obtain the final CER.
Since the results of each of the five folds vary, we compute the average to compare the outcomes of the different network architectures, books, and number of lines.

In each iteration during training one line is randomly chosen out of the data fold.
The number of iterations during training is chosen as 10,000, 12,000, 15,000, 20,0000, 25,000, and 30,000 for the training set size of 60, 100, 150, 250, 500, and 1000, respectively.
These values are fixed for both the default OCRopus and the deep networks.

\subsection{Voting}
\label{sec:voting}
In order to further improve the predictions, we use our Python implementation of ISRI's sequence voting tool to vote on the label sequences that are produced by the CTC-Greedy-Decoder for each fold.
The idea of this voting is to use the faulty sequences of different models to estimate a better one.
Lets assume the following predictions for only three folds in Table \ref{tab:example_voting}.
\begin{table}[t]
    \centering
    \caption{An example for the improvement by using a voting mechanism. Here only three folds are used to derive a result with one instead of two errors, respectively.}
    \label{tab:example_voting}
    \begin{tabular}{rl}
        \hline
               & \textbf{Prediction} \\
        \hline
        Fold 1 & \texttt{An example senience with erors} \\
        Fold 2 & \texttt{A example sentence with erors} \\
        Fold 3 & \texttt{An example entence with error} \\
        \hline 
        Voted & \texttt{An example sentence with erors} \\
        \hline
    \end{tabular}
\end{table}
In a first step, all sentences are aligned as far as possible.
Afterwards all differences are searched and counted, whereby the character with the highest count is kept and the final sequence is computed.
Obviously, there can still be mistakes as seen in the example, but the overall result is improved significantly.

An important prerequisite for the models that are used to vote is that they are almost equally performant, but different in their predictions.
As shown on the examples, errors that occur randomly in one or another sentence can easily be corrected, whereas similar errors such as the forgotten repetition of \texttt{r} can not be found.

The number of models that are used for voting in our experiments is set to 5.
Those are the individual models produced by the 5-fold Cross-Validation on each training set.
A higher amount of models to vote is expected to yield better results, but only if their results are independent.
However, experiments show that all models trained on different folds share a high amount of similar errors, e. g. inserted or deleted spaces, which is why a fold size of 5 is reasonable.

\section{Experiments}

In the following, we present our findings regarding the improvements of deeper architectures and the usage of voting.
Additionally, we also compare the time required for training and prediction of the different architectures, and extend the training corpus to a size of up to 3000 lines to investigate the behaviour of the deep models on many lines.

\subsection{Results of the different Networks}

\begin{figure}
    \centering
    \includegraphics[width=0.9\linewidth]{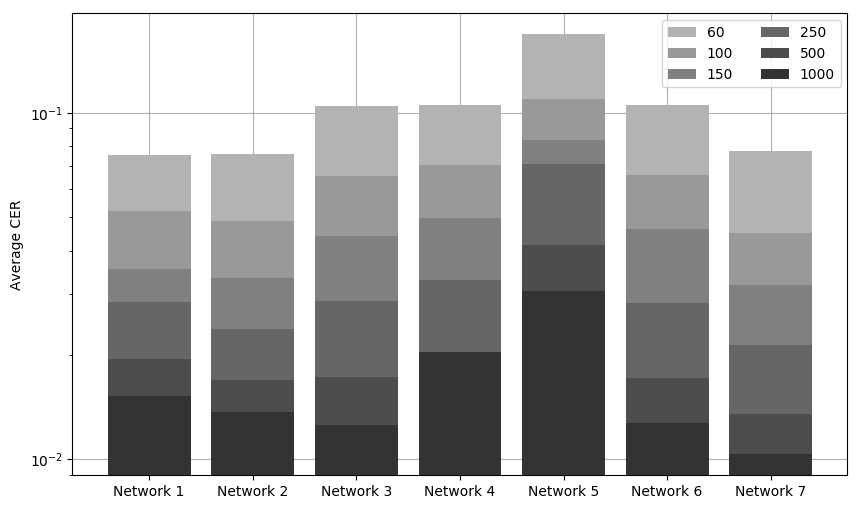}
    \caption{This plot compares the average of the CER on the three books for all seven different network architectures on a logarithmic scale. The shading indicates the average CER for each number of lines used for the 5-cross-fold training.}
    \label{fig:compare_all_models_avg}
\end{figure}

The results obtained by applying all considered network architectures on the three used books are plotted in Figure \ref{fig:compare_all_models_avg}.
Each single bar shows the average fold accuracies for a single book obtained for 60, 100, 150, 200, 500, and 1000 lines in different gray shadings.
The individual fold accuracies for Network 1 and Network 7 are shown in Table \ref{tab:results}.
As expected the overall CER on all experiments is decreased with an increasing number of lines.
The standard OCRopus implementation, Network 1, always yields good and even best results for a low number of lines in the training data set.
Network 2 introduces a single CNN-Pooling layer. Its performance is similar to Network 1 on 60 lines but better on an increasing number of lines.
A second pair of CNN and Pooling, Network 3, increases the gap between high performance on many lines and worse performance on a few lines.
From 250 lines on this network outperforms Network 1 and 1000 lines are required to yield better results than the simpler Network 2.
Network 4 and 5, both designed to allow for an easier prediction of repeated characters, have very high CER even for a high number of lines.
The performance of Network 6 is very similar compared to Network 3, that uses dropout, but the same net structure.
The dropout with the aim to avoid overfitting neither does improve the results for a small number of lines, nor a high number of lines.
However, Network 7 significantly decreases the CER, which is explained by the doubled amount of nodes in the hidden LSTM layer.
This network outperforms all other models starting from 150, which is why this network is used for the further evaluation tasks.

Table \ref{tab:results} shows exemplary the CER for each of the five folds on the three books for 150, 250 and 1000 lines.
The average of these models is used to compute the average improvement of the best network, that is Network 7 to Network 1.
Note, that in practice the individual results of the model for each fold must be combined e.g. by voting (see Section \ref{sec:voting}) in order to obtain a usable result.
Without voting, only one fold could be used to predict a sequence.

The results of the individual folds show no obvious correlation between CERs on the same fold, that is the same training data, and its CERs on different network architectures.
For example the worst fold of the default network on book 1476 using 60 lines is the third fold with an CER of 9.23\%, however this is the second best model for the deep network.
The same can be observed for book 1505 for the same fold: The best outcome of the default network with CER of 5.87\% is the second worst fold for the deep network.

\begin{table}[tp]
    \setlength{\tabcolsep}{0.5em}
    \centering
    \caption{This table shows improvement of a deep CNN-LSTM (Network 7) compared to the default architecture (Network 1) exemplary for 150, 250, and 1000 lines in the training dataset.
    Both the individual results of each Cross-Fold plus their average, and the voting improvements are completely listed.
    The last columns of the fold average and the voting average state the relative improvements of the deep network architecture.
    All numbers are given in percent.}
    \label{tab:results}
    \begin{tabular}{c|l|ccccc|cc|cc}
        \hline
        \multicolumn{11}{c}{\textbf{150 Lines}} \\
        \hline
        \hline
             &       &\multicolumn{5}{c|}{\textbf{CERs per fold}} & \multicolumn{2}{c}{\textbf{Avg.}} & \multicolumn{2}{c}{\textbf{Seq. Voted}} \\
        \hline
        \textbf{Book} & \textbf{Network} & \textbf{1} & \textbf{2} & \textbf{3} & \textbf{4} & \textbf{5} & \textbf{CER} & \textbf{Imp.} & \textbf{CER} & \textbf{Imp.} \\
        \hline
        1476 & Network 1 & 3.87 & 4.14 & 3.64 & 4.30 & 4.53 & 4.01 &      & 2.28 &       \\
             & Network 7 & 3.17 & 3.46 & 3.77 & 3.73 & 3.45 & 3.51 & 12.5 & 2.56 & -12.3 \\
        1488 & Network 1 & 3.01 & 2.70 & 2.92 & 2.96 & 3.23 & 2.96 &      & 1.89 &       \\
             & Network 7 & 2.54 & 2.54 & 2.76 & 2.59 & 3.11 & 2.71 &  8.4 & 2.28 & -20.6 \\
        1505 & Network 1 & 3.67 & 3.74 & 3.30 & 3.51 & 3.49 & 3.54 &      & 2.56 &       \\
             & Network 7 & 2.92 & 3.43 & 3.28 & 3.46 & 3.54 & 3.33 &  5.9 & 2.81 &  -9.8 \\
        \hline
        \textbf{Avg.} &         &      &      &      &      &      &      &  8.9 &      & -14.2 \\
        \hline
        \hline
        \multicolumn{11}{c}{\textbf{250 Lines}} \\
        \hline
        \hline
             &       &\multicolumn{5}{c|}{\textbf{CERs per fold}} & \multicolumn{2}{c}{\textbf{Avg.}} & \multicolumn{2}{c}{\textbf{Seq. Voted}} \\
        \hline
        \textbf{Book} & \textbf{Network} & \textbf{1} & \textbf{2} & \textbf{3} & \textbf{4} & \textbf{5} & \textbf{CER} & \textbf{Imp.} & \textbf{CER} & \textbf{Imp.} \\
        \hline
        1476 & Network 1 & 3.03 & 3.06 & 2.93 & 4.21 & 2.99 & 3.24 &      & 1.93 &       \\
             & Network 7 & 2.07 & 2.54 & 2.25 & 2.29 & 2.12 & 2.26 & 30.2 & 1.75 &   9.3 \\
        1488 & Network 1 & 2.94 & 2.30 & 2.00 & 3.40 & 2.32 & 2.59 &      & 1.50 &       \\
             & Network 7 & 1.85 & 1.63 & 1.89 & 1.77 & 1.93 & 1.81 & 30.1 & 1.29 &  14.0 \\
        1505 & Network 1 & 2.68 & 2.70 & 2.78 & 2.75 & 2.63 & 2.71 &      & 1.87 &       \\
             & Network 7 & 2.35 & 2.33 & 2.22 & 2.42 & 2.35 & 2.33 & 14.0 & 1.91 &  -2.1 \\
        \hline
        \textbf{Avg.} &         &      &      &      &      &      &      & 25.8 &      &   7.1 \\
        \hline
        \hline
        \multicolumn{11}{c}{\textbf{1000 Lines}} \\
        \hline
        \hline
             &       &\multicolumn{5}{c|}{\textbf{CERs per fold}} & \multicolumn{2}{c}{\textbf{Avg.}} & \multicolumn{2}{c}{\textbf{Seq. Voted}} \\
        \hline
        \textbf{Book} & \textbf{Network} & \textbf{1} & \textbf{2} & \textbf{3} & \textbf{4} & \textbf{5} & \textbf{CER} & \textbf{Imp.} & \textbf{CER} & \textbf{Imp.} \\
        \hline
        1476 & Network 1 & 1.46 & 1.52 & 1.52 & 1.56 & 1.68 & 1.55 &      & 1.11 &       \\
             & Network 7 & 0.85 & 0.99 & 0.86 & 0.84 & 0.78 & 0.86 & 44.4 & 0.65 &  41.1 \\
        1488 & Network 1 & 1.15 & 1.20 & 1.08 & 1.03 & 1.38 & 1.17 &      & 0.77 &       \\
             & Network 7 & 0.71 & 0.69 & 0.65 & 0.77 & 0.77 & 0.72 & 38.5 & 0.50 &  35.9 \\
        1505 & Network 1 & 1.96 & 1.87 & 1.77 & 1.85 & 1.77 & 1.84 &      & 1.44 &       \\
             & Network 7 & 1.42 & 1.59 & 1.58 & 1.60 & 1.41 & 1.52 & 17.4 & 1.23 &  14.6 \\
        \hline
        \textbf{Avg.} &         &      &      &      &      &      &      & 33.4 &      &  30.5 \\
        \hline
        
    \end{tabular}
\end{table}

The relative improvements that are exemplary shown in Table \ref{tab:results} are listed for all the different number of lines in Table \ref{tab:improvements} for Network 1 and Network 7.
In the left section the relative improvements of the average of the folds is shown, on the right hand side, the improvement when using voting (see Section \ref{sec:evaluation_of_voting}).
The deeper network architectures yield an increasingly better average CER from 100 lines onward.
On 1000 lines the best improvement on a single book is 44.4\% and the average over all books is increased by 33.3\%.

\begin{figure}[t]
    \centering
    \includegraphics[width=0.9\linewidth]{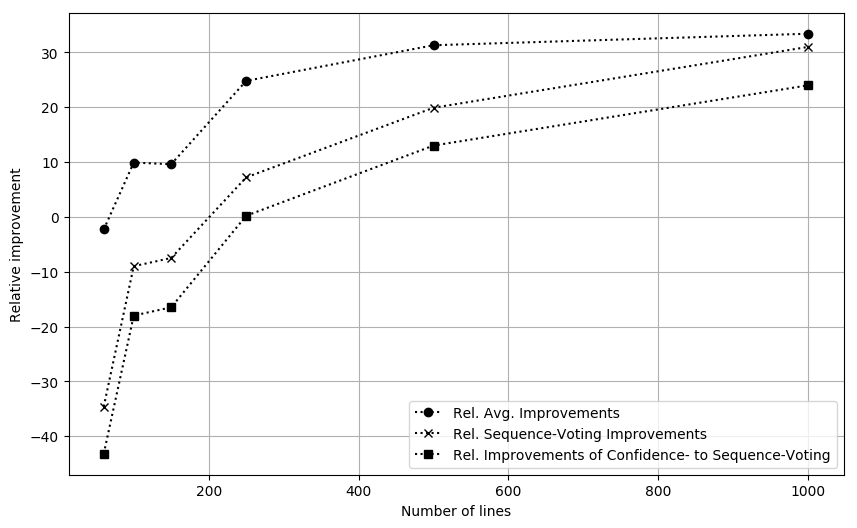}
    \caption{Relative averaged improvements of a deep network versus the default OCRopus network. The solid points indicate the relative improvements based on the averages of the Cross-Fold results.
    The crosses show the relative improvements using sequence voting.
    The squares indicate the relative improvements when using confidence voting on the default OCRopus network and sequence voting on the deep network.}
    \label{fig:relative_improvements}
\end{figure}

A plot of the relative increase dependent on the number of lines is shown in Figure \ref{fig:relative_improvements} (solid points).
The increasing slope is almost flat at the right most point which indicates that more lines used for training a deep model will still yield a better model compared to the default model, but its relative improvement is eventually constant.
An even deeper network might increase the relative improvement if more lines are available.

\begin{table}[t]
    \centering
    \caption{Relative improvement of the deep CNN (Network 7) compared to default OCRopus (Network 1) listed for all three books and the six variations of training data amount.
    The last column shows the relative improvement when using sequence voting on Network 7 and the more sophisticated confidence voting on Network 1.
    All numbers are given in percent.}
    \label{tab:improvements}
    \begin{tabular}{r|ccc|c|ccc|c||c}
        \hline
              & \multicolumn{4}{c|}{\textbf{Improvement over folds}} & \multicolumn{4}{c||}{\textbf{Improvement over voting}} & \textbf{Conf.} \\
              & \multicolumn{3}{c|}{\textbf{Book}} & & \multicolumn{3}{c|}{\textbf{Book}} & & \textbf{Voted} \\ 
        \textbf{Lines} & \textbf{1476} & \textbf{1488} & \textbf{1505} & \textbf{Avg.} & \textbf{1476} & \textbf{1488} & \textbf{1505} & \textbf{Avg.} & \textbf{Avg.} \\
        \hline
           60 & -7.9 & -3.7 &  4.8 & -2.3 & -32.7 & -49.5 & -21.7 & -34.6 & -43.2 \\
          100 & 30.7 & 1.1  & -2.2 &  9.9 & -11.9 &  -5.7 &  -9.5 &  -9.0 & -18.0 \\
          150 & 14.2 & 8.7  &  6.0 &  9.6 &   7.7 & -20.3 &  -9.9 &  -7.5 & -16.5 \\
          250 & 30.5 & 30.1 & 13.8 & 24.8 &   9.3 &  14.4 &  -2.2 &   7.2 &   0.2 \\
          500 & 38.6 & 34.5 & 20.9 & 31.3 &  15.3 &  33.8 &  10.6 &  19.9 &  13.0 \\
         1000 & 44.3 & 38.4 & 17.6 & 33.4 &  42.3 &  35.9 &  14.8 &  31.0 &  24.0 \\
         \hline
    \end{tabular}
\end{table}

\subsection{Evaluation of voting}
\label{sec:evaluation_of_voting}

\begin{figure}[t]
    \centering
    \includegraphics[width=0.9\linewidth]{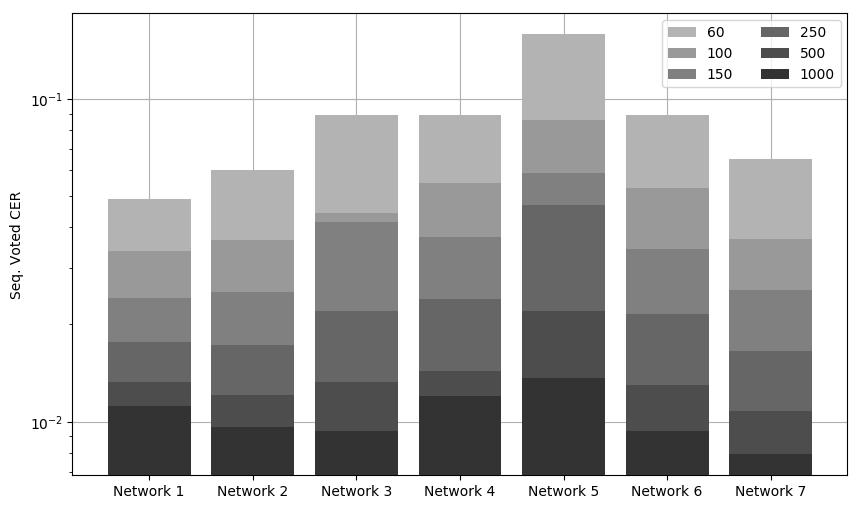}
    \caption{This plot compares the average of the CER on the three books for all seven different network architectures using voting on a logarithmic scale. The shading indicates the average CER for each number of lines used for the 5-cross-fold training using voting.}
    \label{fig:compare_all_models_voted}
\end{figure}

The average error rates based on applying the sequence voting algorithm are shown in Table \ref{tab:results} and Figure \ref{fig:compare_all_models_voted}.
As expected, voting improves the accuracy on all experiments by a significant amount.
The voted model even reaches a best value of below $0.5\%$ CER on the book 1488.
However, the default Network 1 benefits by a higher margin than the deep Network 7.
In Table \ref{tab:results} Network 7 is still worse than Network 1 at 150 lines, starting from 250 lines the CER is reduced.

The overall shape of Network 2 to 7 in Figure \ref{fig:compare_all_models_voted} is similar to the one in Figure \ref{fig:compare_all_models_avg} that shows the averages without voting, but Network 1 is clearly shrunken, particularly for 60 lines.
The relative improvements shown in Table \ref{tab:improvements} and Figure \ref{fig:relative_improvements} of Network 7 to Network 1 using voting clarify this behaviour.
The deep network architecture performs significantly worse than the default network by an average factor of $-31.5\%$ when using 60 lines, however it has a similar improvement of $>30\%$ compared to the average if 1000 lines are used.
Considering the slope of Figure \ref{fig:relative_improvements} it is to be expected that the relative improvement approaches a limit value.
Hence, the used deep model is expected to be still better than the default model by absolute values, but the relative gap appears to be constant.
One approach is to use even deeper architectures if more lines are available for training of more complex models.

As shown, voting has a higher impact on the default OCRopus network than on the used deep network, especially if only a few lines are used for training.
Thus, the individual models must be more diverse to allow for a more effective voting.
The evaluation of errors (see Table \ref{tab:confusion_errors} in Section \ref{sec:evaluation_of_errors}) shows that besides insertion and deletions of spaces, the errors of deep networks are mostly missing characters, while the errors of the default network are mostly confusions of e.g. e$\leftrightarrow$c or n$\leftrightarrow$r in both directions
Therefore, voting of deep models that all omit single character predictions (compare Table \ref{tab:confusion_errors}) and therefore suffer from similar errors can not benefit by such a high amount compared to the default models whose errors are more random.

Finally, the last column in Table \ref{tab:improvements} shows the relative improvements of Network 7 compared to Network 1 when using sequence voting for Network 7 and the more sophisticated confidence voting for Network 1 proposed by \cite{reul2017voting}.
The expected benefit of confidence voting of 5\% to 10\% compared to sequence voting is clearly visible.
The squared curve in Figure \ref{fig:relative_improvements} visualizes this behaviour by an almost constant shift compared to using sequence voting in both networks.
Consequently, confidence voting must be considered for further improvements of the deep networks.

\subsection{Evaluation of errors}
\label{sec:evaluation_of_errors}

Table \ref{tab:confusion_errors} lists the ten most common errors of Network 7 after voting when predicting the evaluation dataset of 1476.
The results are shown for 150 lines and 1000 lines on the left and right columns, respectively.
Both networks most commonly do wrong insertions of spaces, however the model trained with 150 lines then mostly misses the prediction of characters.
This is the reason why we implemented Networks 4 and 5, since they have a higher chance to predict characters, however the overall results show that these models perform worse.
\begin{table}[t]
    \centering
    \caption{The 10 most common errors made by Network 7 after voting on the evaluation dataset of book 1476. The left results are for 150 training lines, the right results for 1000 training lines. An underscore represents deletions or insertions of characters.}
    \label{tab:confusion_errors}
    \begin{tabular}{r|cc||r|cc}
        \hline
        \multicolumn{3}{c||}{\textbf{150 Lines}} & \multicolumn{3}{c}{\textbf{1000 Lines}} \\
        \textbf{Count} & \textbf{Predicted} & \textbf{True} & \textbf{Count} & \textbf{Predicted} & \textbf{True} \\
        \hline
        62 & \_ & SPACE & 36 & SPACE & \_ \\
        53 & SPACE & \_ & 29 & \_ & SPACE \\
        45 & \_ & i &     10 & \_ & i \\
        32 & \_ & n &      6 & t & r \\
        20 & \_ & r &      4 & \longs & f\\
        18 & \_ & o &      4 & \_ & n \\
        17 & \_ & t &      3 & t & \_ \\
        13 & \_ & v &      3 & i & \_ \\
        10 & \_ & s &      2 & w\_ & vo \\
         9 & \_ & d &      2 & f & \longs \\
        \hline
    \end{tabular}
\end{table}
Network 7 trained with 1000 lines shows errors among the top ten that are expected, e.g. wrong classifications of f and \longs{} (old German long s).

\subsection{Evaluation of time}

In this section we compare the training time and the time for the prediction of a complete book of the default OCRopus network (1) and our best performing deep network (7).
All times were measured on an Intel Core i7-5820K processor using 1, 2, 4 or 8 threads for each experiment.
During training the parallelism is internally used in Numpy for the default OCRopus implementation and in the TensorFlow-Backend for our deep networks.
For predicting the default OCRopus implementation copies the individual model for each thread and uses only one thread in the internal operations of Numpy.
Our TensorFlow implementation instead creates only one model and predicts one batch consisting of 20 lines using all desired threads.
Note that each line has to be processed five times during the prediction because the voting algorithm.
Table \ref{tab:timing} reports our findings for the averages across the three used books.

\begin{table}[t]
    \centering
    \caption{Average times for training and prediction of a single line for all three books.
    Note that during prediction each line has to be processed five times due to the voting of the five folds.
    The timing procedure was conducted for a various number of threads.}
    \label{tab:timing}
    \begin{tabular}{r|cccc|cccc}
        \hline
                         & \multicolumn{4}{c|}{\textbf{Training in seconds}} & \multicolumn{4}{c}{\textbf{Prediction in seconds}} \\
        \textbf{Threads} & 1    & 2    & 4    & 8    & 1   & 2   & 4   & 8 \\
        \hline
        \textbf{Network 1} & 0.28 & 0.27 & 0.30 & 0.40 & 0.89 & 0.48 & 0.25 & 0.16 \\
        \textbf{Network 7} & 0.40 & 0.30 & 0.25 & 0.26 & 0.51 & 0.31 & 0.20 & 0.14 \\
        \hline
    \end{tabular}
\end{table}

First of all, the results for the training time show that even if the deeper network consists of way more parameters and more operations, the optimized TensorFlow implementation is only slightly slower than the default implementation based on Numpy, when only one thread is used.
However, the shallow default LSTM net can not benefit from a higher number of threads, instead it even suffers from a too high count.
The reason is, that the underlying matrix multiplications are too low dimensional in order to be relevant for multiprocessing.
As expected, the deep networks benefit up to 4 threads, as the training time is decreased by an average factor of approximately 40\%.
The reason is, that the convolution operations that are carried out on each pixel on the full image profit from a parallel computation.
As a result, the deep network is faster than the default implementation when allowing several cores during the training.
Our results show, that a number of 4 is sufficient.

The number of processed lines per second is massively increased by using batch training on a GPU as reported by \cite{breuel17hybridCNN-LSTM}.
However, the question whether this improves the accuracy or at least leads to a decrease in absolute training time has not been reported yet and is not in the focus of this paper.

The average time required to predict one line for the three books shows that the deep network requires almost half the time compared to the default OCRopus implementation, whereby the time for voting can be neglected.
An explanation is the native C++ implementation of the CTC-Decoder-Algorithm compared to the slow Python implementation of OCRopus.
Using a higher number of threads the required time for prediction almost shrinks linearly for the default implementation, since each thread computes a single independent model by construction.
The TensorFlow implementation still benefits from a higher thread count, but by a reduced factor due to the core sharing of batch versus convolutional operation.
Yet, for all the tested thread counts the TensorFlow implementation is a bit faster than default OCRopus.
Obviously, the prediction time can be further reduced by the usage of a GPU due to the high throughput of images per second.

\subsection{Increasing the training data above 1000}
The available amount data for the three books allows us to increase the training set size up to 2000, 3000, and 2000 for the books 1476, 1488, and 1505, respectively (compare Table \ref{tab:books}).
The averaged CER of the folds and after voting is shown in Table \ref{tab:huge_lines} by usage of Network 7.
\begin{table}[t]
    \centering
    \caption{Decrease of the CER for using more than 1000 lines for training Network 7. All values are given in percent.}
    \label{tab:huge_lines}
    \begin{tabular}{r|ccc|c|ccc|c|}
        \hline
              & \multicolumn{4}{c|}{\textbf{Averaged CER of folds}} & \multicolumn{4}{c|}{\textbf{CER using voting}} \\
              & \multicolumn{3}{c|}{\textbf{Book}} & & \multicolumn{3}{c|}{\textbf{Book}} & \\ 
        \textbf{Lines} & \textbf{1476} & \textbf{1488} & \textbf{1505} & \textbf{Avg.} & \textbf{1476} & \textbf{1488} & \textbf{1505} & \textbf{Avg.} \\
        \hline
        1000 & 0.86 & 0.72 & 1.5 & 1.0  & 0.65 & 0.50 & 1.2 & 0.78 \\
        1500 & 0.79 & 0.62 & 1.4 & 0.94 & 0.55 & 0.43 & 1.1 & 0.69 \\
        2000 & 0.76 & 0.61 & 1.3 & 0.89 & 0.55 & 0.43 & 1.0 & 0.66 \\
        3000 &  $-$ & 0.50 & $-$ &  $-$ &  $-$ & 0.37 & $-$ &  $-$ \\
        \hline
    \end{tabular}
\end{table}
As expected, the use of even more lines for training further shrinks the CER in all experiments with and without voting by a significant amount.
Thus, we reached a maximum CER of 1.0\% after voting on all three books.

\section{Conclusion and future work}
In this paper we introduced combinations of CNN- and LSTM-Networks to push the error rate to a minimum achieving best values below $1\%$ CER and a relative improvement compared to standard models of above $40\%$.
The enhancements are massively increased by a larger amount of available training data and the introduction of voting mechanisms.
Although the proposed deeper networks have a higher number of parameters the absolute training and prediction time is in the same order of magnitude compared to the standard model.

For further improvements, we propose transfer learning by using a pretrained model as initial instance and a following finetuning on a specific book.
Using several datasets for pretraining different models should affect beneficially the voting algorithms because many independent models are available.
Similarly, different voters could be created by variations of the network architectures.
Moreover, more sophisticated voting algorithms such as the proposed confidence voting of \cite{reul2017voting} that includes the output probabilities of each possible character in the voting compared to a simple majority vote, might further decrease the CER.

To gather more data for training the deep networks it is expected that synthetic data or data augmentation leads to significant improvements.
If more data is available even deeper architectures must be considered.
An implementation of the training on a GPU combined with batch-wise training might reduce the absolute training and prediction time of a single model due to a higher throughput of lines per second.

In summary, it can be stated that the application of deep CNN-LSTM-networks open the doors to very promising approach to establish a net benchmark for OCR of early printed books and despite the historical focus of this paper also on any other print.

\bibliographystyle{apa}
{\small \bibliography{main}}

\end{document}